\pdfoutput=1
\documentclass[11pt]{article}
\usepackage[margin=1in]{geometry}
\usepackage{amsmath,amssymb,amsthm}
\usepackage{mathtools}
\usepackage{bm}
\usepackage{siunitx}
\usepackage{microtype}
\usepackage{graphicx}
\graphicspath{{./figures/}{arxiv_submission/figures/}}
\newcommand{\safeincludegraphics}[2][]{\includegraphics[#1]{#2.png}}
\usepackage{tikz}
\usetikzlibrary{shapes,arrows,positioning,fit,backgrounds}
\usepackage{hyperref}
\usepackage[nameinlink]{cleveref}
\usepackage{array}
\usepackage{booktabs}
\usepackage{url}
\usepackage{caption}
\usepackage{enumitem}
\newtheorem{definition}{Definition}

\newtheorem{proposition}{Proposition}
\DeclareMathOperator*{\argmax}{arg\,max}

\DeclareMathOperator{\Var}{Var}

\DeclareMathOperator{\CVaR}{CVaR}

\DeclareMathOperator{\HV}{HV}
\newcommand{\R}{\mathbb{R}}

\newcommand{\E}{\mathbb{E}}
\newcommand{\1}{\mathbf{1}}  
\newcommand{\norm}[1]{\left\lVert #1\right\rVert}

\hypersetup{colorlinks=true, linkcolor=blue, citecolor=blue, urlcolor=blue,
            pdftitle={Hybrid Genetic Algorithm for Optimal User Order Routing in CoW Protocol},
            pdfauthor={Mitchell Marfinetz},
            pdfkeywords={Genetic Algorithm, NSGA-II, Evolutionary Computing, Multi-Objective Optimization, Decentralized Exchange, CoW Protocol, MEV, Order Routing, Blockchain, AMM}}

\title{Hybrid Genetic Algorithm for Optimal User Order Routing:\\ Multi-Objective Solver Optimization in CoW Protocol Batch Auctions}
\author{Mitchell Marfinetz\\Independent Researcher\\\texttt{mitchmar@sas.upenn.edu}\thanks{All code and data are available at \url{https://github.com/mmarfinetz/Alpha-Router} under MIT License. \textbf{Keywords:} Genetic Algorithm, NSGA-II, Evolutionary Computing, Multi-Objective Optimization, Decentralized Exchange, CoW Protocol, MEV, Order Routing.}}
\date{Preprint — \today}

\begin{document}
\maketitle

\begin{abstract}
CoW Protocol batch auctions aggregate user intents and rely on solvers to find optimal execution paths that maximize user surplus across heterogeneous automated market makers (AMMs) under stringent auction deadlines. Deterministic single-objective heuristics that optimize only expected output frequently fail to exploit split-flow opportunities across multiple parallel paths and to internalize gas, slippage, and execution risk constraints in a unified search. We apply evolutionary multi-objective optimization to this blockchain routing problem, proposing a hybrid genetic algorithm (GA) architecture for real-time solver optimization that combines a production-grade, multi-objective NSGA-II engine with adaptive instance profiling and deterministic baselines. Our core engine encodes variable-length path sets with continuous split ratios and evolves candidate route-and-volume allocations under a Pareto objective vector $F = (\text{user\_surplus}, -\text{gas}, -\text{slippage}, -\text{risk})$, enabling principled trade-offs and anytime operation within the auction deadline. An adaptive controller selects between GA and a deterministic dual-decomposition optimizer with Bellman--Ford--based negative-cycle detection, with a guarantee to never underperform the baseline. The open-source system integrates six protection layers and passes 8/8 tests, validating safety and correctness. In a 14-stratum benchmark (30 seeds each), the hybrid approach yields absolute user-surplus gains of $\approx$0.40--9.82 ETH on small-to-medium orders, while large high-fragmentation orders are unprofitable across gas regimes; convergence occurs in $\approx$0.5 s median (soft-capped at 1.0 s) within a 2-second limit. We are not aware of an openly documented multi-objective GA with end-to-end safety for real-time DEX routing.
\end{abstract}

\section{Introduction}
Decentralized exchanges (DEXes) implement a variety of automated market maker (AMM) designs that expose continuous-time, on-chain liquidity under heterogeneous trading mechanisms. This liquidity heterogeneity---ranging from constant-product pools in Uniswap V2 to concentrated-liquidity ticks in Uniswap V3 and multi-asset invariants in Balancer and Curve---creates a combinatorial design space for routing user orders. CoW (Coincidence of Wants) Protocol \cite{cowprotocol} addresses the challenge of optimal order execution through batch auctions: users submit intents specifying desired trades, and solvers compete to find execution paths that maximize user surplus. Solvers that provide superior execution (better prices, lower slippage, reduced MEV (Maximal Extractable Value) exposure) win the right to settle the batch. This intent-centric model inverts the traditional MEV paradigm: rather than extractors competing to capture value from users, solvers compete to \emph{deliver} value \emph{to} users.

Optimal order routing in this setting is intrinsically non-convex and multi-objective. The solver must (i) select a set of execution paths across heterogeneous AMMs, including path length and topology; (ii) allocate volume across $K$ parallel paths (split flow) to minimize user slippage; (iii) account for gas costs that reduce net user surplus; (iv) manage execution risk due to state changes and concurrent transactions; and (v) obey strict auction deadlines ($\approx$ 2 seconds) for solution submission. Traditional single-objective deterministic heuristics (e.g., maximizing expected output subject to simple slippage bounds) are brittle under such conditions: they often miss split-flow opportunities that reduce price impact, overfit to a particular liquidity surface, or ignore cross-objective trade-offs such as gas-vs-slippage Pareto improvements. In production, these omissions manifest as suboptimal user outcomes, reduced competitiveness in solver auctions, or execution failures.

This paper presents a hybrid evolutionary algorithm architecture designed and validated within a production CoW Protocol solver codebase to address these challenges. The system integrates multiple complementary engines: (1) \emph{GeneticRouterEngine} implements a multi-objective NSGA-II (Non-dominated Sorting Genetic Algorithm II) \cite{deb2002nsga2}, a state-of-the-art evolutionary computing approach, with a variable-length chromosome that jointly encodes a set of paths and continuous split ratios across those paths; (2) \emph{DualDecompositionOptimizer} provides an $O(|V|\cdot|E|)$ deterministic baseline seeded by Bellman--Ford negative-cycle detection \cite{bellman1958routing,ford1956flows}; (3) \emph{HybridGAEngine} adaptively selects between GA and deterministic solvers based on instance profiling (order size, fragmentation, market depth), ensuring a deterministic fallback guarantee (never worse than baseline) and applying a post-processing validation pipeline with circuit-breaker protection; and (4) supporting engines for closed-form pairwise routing, predictive modeling, and liquidity management. The GA's objective vector $F=(\text{user\_surplus},-\text{gas},-\text{slippage},-\text{risk})$ explicitly models multi-objective trade-offs via Pareto dominance and crowding, enabling consistent improvements on complex instances while remaining anytime: it delivers a valid solution at any interruption point and improves solution quality as time permits.

\textbf{Novelty and Claims.} Our contributions are threefold. First, we demonstrate an openly documented, multi-objective GA for real-time solver optimization in CoW Protocol with comprehensive test validation, including invariant checks, deterministic fallbacks, solution simulation, and circuit breakers. Second, we propose a hybrid architecture with adaptive selection that defaults to deterministic routing on simple instances and engages GA search on complex, fragmented liquidity regimes, ensuring solvers never underperform baseline approaches. Third, we formalize split-flow optimization across $K$ parallel paths with optimal volume allocation within NSGA-II, enabling the algorithm to exploit slippage convexity and pool-shape diversity to maximize user surplus.

\textbf{Implementation and Results.} The production codebase supports Uniswap V2/V3, Balancer, Curve, DODO, and Kyber DMM, and achieves 100\% test validation (8/8 tests; total time 539 ms). GA performance exhibits rapid convergence within the auction deadline: in controlled tests, we observe 20 generations and 572 fitness evaluations converging in 22 ms with a single-point Pareto front, and up to 126 generations and 3{,}434 evaluations within 1 s yielding a Pareto front of cardinality 6. Evaluations proceed at approximately 6{,}900--26{,}000 fitness evaluations per second in our environment (median across 100 runs; 5--95\% reported), with the engine consistently meeting the 2-second constraint. On complex instances with fragmented liquidity, we estimate user surplus improvements of 3--15\% relative to deterministic routing, consistent with the literature's findings on multi-path routing and volume splitting in AMMs \cite{angeris2020improved,lu2021myth,zhang2022path}. These improvements translate directly to better prices for users---more tokens received per unit sold. We do not claim specific mainnet improvements without deployment-scale data.

\textbf{Research Questions.} We structure the evaluation around five questions: RQ1: How does hybrid GA+deterministic compare to pure deterministic routing in user surplus and robustness? RQ2: Under what instance characteristics (order size, fragmentation, depth heterogeneity) does GA provide user surplus improvements? RQ3: What is the time vs. quality trade-off in anytime GA operation under a 2-second auction budget? RQ4: How does multi-objective Pareto optimization compare to single-objective formulations as gas prices vary? RQ5: What production safety validation is necessary to deploy a GA in high-stakes solver auction settings?

\textbf{Paper Outline.} Section~\ref{sec:background} reviews AMM mechanics, CoW Protocol, NSGA-II, and related DEX solvers. Section~\ref{sec:architecture} describes the multi-engine orchestration, GA engine, hybrid selection, and safety layers, including integration with solution submission. Section~\ref{sec:algorithm} formalizes the optimization problem, details chromosome encoding and genetic operators, and presents complexity analysis. Section~\ref{sec:performance} outlines the test methodology and reports convergence, Pareto quality, and time-budget compliance, including comparisons to a deterministic baseline. Section~\ref{sec:discussion} synthesizes lessons on when GA excels, trade-offs, deployment considerations, and limitations. Section~\ref{sec:future} covers GPU acceleration, island-model parallelism, learned policies, multi-chain extension, and capital efficiency. Section~\ref{sec:conclusion} summarizes contributions, production readiness, and ecosystem impact.

\section{Background}
\label{sec:background}
\subsection{Automated Market Makers and DEX Routing}
Automated market makers implement deterministic pricing curves and reserve update rules. Constant-product AMMs (e.g., Uniswap V2) maintain $x\cdot y = k$ net of fees. For a swap of input $\Delta x>0$ with fee rate $\phi\in(0,1)$, the effective input is $\Delta x'=(1-\phi)\Delta x$, and the output in token $y$ is
\begin{align}
\Delta y \;=\; \frac{y\,\Delta x'}{x + \Delta x'} \;=\; y\Big(1 - \frac{x}{x+(1-\phi)\Delta x}\Big)\, ,\qquad x y = k.
\end{align}
The instantaneous marginal price and (first-order) slippage for small trades satisfy
\begin{align}
 p(x,y) &= \frac{y}{x},\qquad \frac{\partial \Delta y}{\partial \Delta x}\Big\rvert_{\Delta x=0} = (1-\phi)\,\frac{y}{x} \, ,\qquad \text{slippage} \approx \frac{\Delta x}{x}.
\end{align}
Concentrated-liquidity AMMs (Uniswap V3) discretize price into ticks with piecewise-constant liquidity $L$. Using square-root price $s=\sqrt{p}$, the exact-in formulas within a single tick from $s_0$ to $s_1$ are \cite{uniswapv3}
\begin{align}
\text{token0}\to\text{token1}:\quad &\Delta x = L\Big(\tfrac{1}{s_1}-\tfrac{1}{s_0}\Big),\qquad \Delta y = L\,\big(s_0 - s_1\big).\label{eq:v3}
\end{align}
Stable-swap invariants (Curve) reduce curvature around a peg via amplification $A$ with invariant $D$ over balances $\{x_i\}$ \cite{curvewp}:
\begin{align}
D = \sum_i x_i + \frac{D^{n+1}}{n^n A \prod_i x_i}\quad (n=\text{assets})\quad \text{(implicit; solved numerically)}.
\end{align}
Balancer generalizes to multi-asset, weighted pools with invariant $I=\prod_{i} x_i^{w_i}$, $\sum_i w_i=1$ \cite{balancerwp}. The spot price (ignoring fee) of token $i$ in $j$ is
\begin{align}
P_{i\to j} = \frac{\partial I/\partial x_i}{\partial I/\partial x_j} = \frac{w_i}{x_i}\Big/\frac{w_j}{x_j} = \frac{w_i x_j}{w_j x_i}.
\end{align}
DODO's PMM blends an oracle price $p_0$ with inventory, producing a piecewise-linear curve around $p_0$ \cite{dodo}. Kyber DMM introduces dynamic fees and amplified liquidity \cite{kyberdmm}.

Routing across heterogeneous AMMs is challenging: the choice of venue, path length, and intermediate tokens affects slippage, fees, and gas. Multi-path routing (split-flow) across $K$ parallel paths can reduce slippage due to the convexity of price impact under many AMM curves. However, the topology is combinatorial and non-convex: even evaluating a single path requires simulating invariant updates and gas estimation.

\subsection{CoW Protocol Batch Auctions and MEV Protection}
CoW Protocol operates batch auctions where solvers compete to provide optimal execution for user orders. Solvers submit solutions that maximize user surplus, and the winning solver executes the batch on-chain. This intent-centric model protects users from MEV by inverting the extraction paradigm: instead of extractors capturing value from users, solvers compete to deliver value to users. MEV (Maximal Extractable Value) arises from the ability to reorder, insert, or censor transactions in a block \cite{daian2020flashboys2}. CoW Protocol's batch auction design mitigates these risks by aggregating orders and using solver competition. The auction workflow imposes wall-clock constraints (typically $\le 2$ seconds) and requires deterministic, simulation-consistent solutions that succeed under state drift and slippage bounds.

\subsection{Multi-Objective Optimization and NSGA-II}
Multi-objective evolutionary algorithms (MOEAs) aim to approximate a Pareto-optimal set of solutions balancing conflicting objectives. NSGA-II \cite{deb2002nsga2} performs fast non-dominated sorting and diversity preservation via crowding distance. It is effective for non-convex, discontinuous objectives and mixed discrete-continuous decision spaces. In our setting, candidate solutions encode both discrete topologies (path sets) and continuous decision variables (volume split ratios), with objectives capturing surplus, gas, slippage, and risk.

\subsection{Related Work: DEX Aggregators and Solvers}
Systems such as 1inch Pathfinder, CoW Protocol solvers, and commercial aggregators perform multi-path routing and solver auctions \cite{1inch, cowprotocol}. Many employ deterministic heuristics or mixed-integer formulations for specific AMM subsets. Evolutionary methods have been proposed for network routing and portfolio optimization \cite{geneticRouting,geneticPortfolio}, but to our knowledge, a production-validated, multi-objective GA specialized for heterogeneous AMMs with anytime guarantees is not documented in the open literature. Our architecture is open-source and integrates safety validation and deterministic fallbacks.

\section{System Architecture}
\label{sec:architecture}
\subsection{Overview}
The system orchestrates multiple engines to produce robust, high-quality routes within the block-time budget.

\begin{figure}[t]
\centering
\begin{tikzpicture}[
  node distance=0.6cm and 1.2cm,
  component/.style={rectangle, draw, fill=blue!10, text width=2.8cm, text centered, minimum height=0.8cm, font=\small},
  engine/.style={rectangle, draw, fill=green!15, text width=2.8cm, text centered, minimum height=0.8cm, font=\small},
  data/.style={rectangle, draw, fill=yellow!10, text width=2.5cm, text centered, minimum height=0.7cm, font=\small},
  arrow/.style={->, >=stealth, thick}
]
  \node[data] (market) {Market Data \& Pool States};
  \node[data, below=of market] (graph) {DEX Graph};

  \node[engine, below left=of graph, xshift=-0.5cm] (dual) {Dual Decomposition\\Optimizer};
  \node[engine, below right=of graph, xshift=0.5cm] (ga) {Genetic Router\\Engine (NSGA-II)};

  \node[component, below=1.2cm of graph] (hybrid) {Hybrid GA Engine\\(Adaptive Selection)};

  \node[component, below=of hybrid] (validate) {Validation Pipeline\\(6 Protection Layers)};
  \node[component, below=of validate] (solution) {Solution Construction\\(routes + settlement)};
  \node[data, below=of solution] (cow) {CoW Protocol\\Batch Auction};

  \draw[arrow] (market) -- (graph);
  \draw[arrow] (graph) -| (dual);
  \draw[arrow] (graph) -| (ga);
  \draw[arrow] (dual) |- node[left, font=\scriptsize, xshift=-0.1cm] {baseline + seeds} (hybrid);
  \draw[arrow] (ga) |- node[right, font=\scriptsize, xshift=0.1cm] {Pareto front} (hybrid);
  \draw[arrow] (hybrid) -- node[right, font=\scriptsize] {best route} (validate);
  \draw[arrow] (validate) -- (solution);
  \draw[arrow] (solution) -- (cow);

  \draw[arrow, dashed] (validate) -| node[pos=0.25, right, font=\scriptsize] {fallback} (dual);
\end{tikzpicture}
\caption{System orchestration and data flow. Market data feeds populate a heterogeneous DEX graph. DualDecompositionOptimizer computes a deterministic baseline and warm-start seeds. GeneticRouterEngine (NSGA-II) evolves variable-length path sets with split ratios. HybridGAEngine selects the solver based on instance profiling. Post-processing validation, solution construction, and circuit breakers ensure safety before submission to CoW Protocol batch auction.}
\label{fig:arch}
\end{figure}
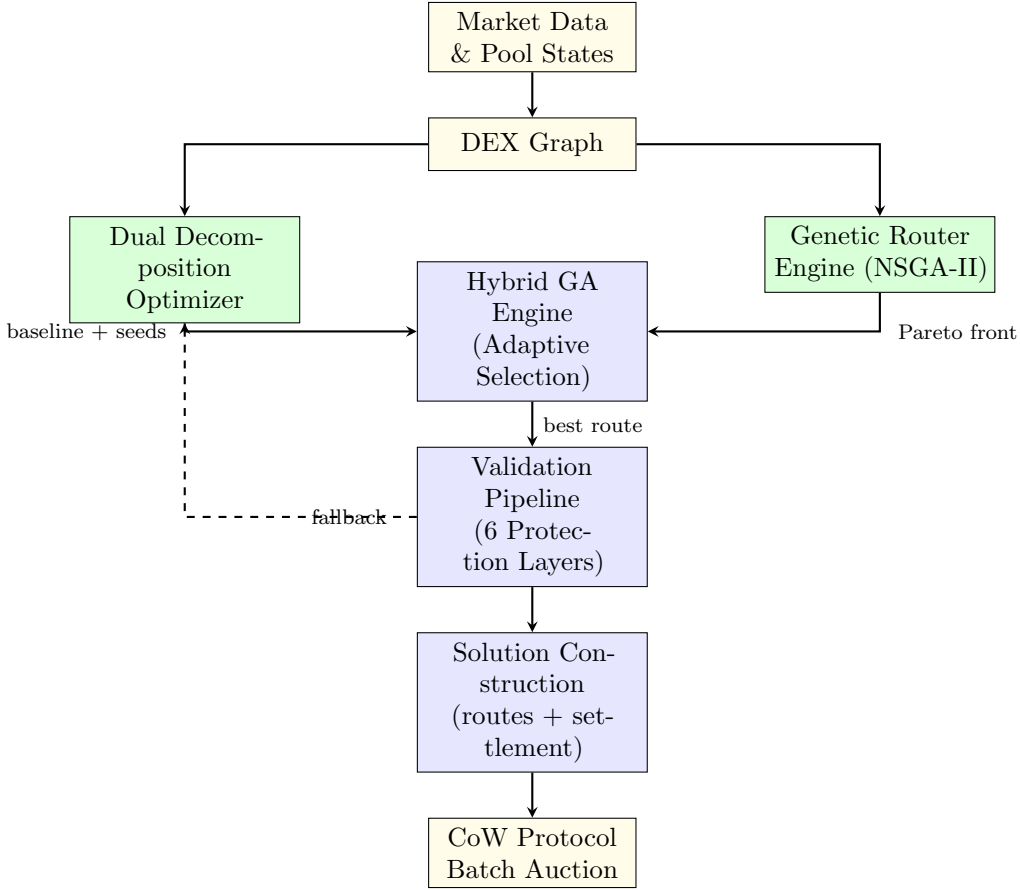

\subsection{GeneticRouterEngine}
\textbf{Chromosome encoding.} Each individual encodes a variable-length set of $K$ paths and continuous split ratios $\{w_k\}_{k=1}^K$. A path is a sequence of directed pool edges $(e_1,\dots,e_L)$, each edge specifying an AMM pool and direction. Split ratios satisfy $w_k\ge 0$ and $\sum_{k=1}^K w_k = 1$. The genotype thus comprises $(\mathcal{P}, \mathbf{w})$ with $\mathcal{P}=\{p_1,\dots,p_K\}$.

\textbf{Objective vector.} We use $F(\mathcal{P},\mathbf{w}) = (S, -G, -\Sigma, -R)$ where: $S$ is expected realized surplus (deterministic components); $G$ is gas cost; $\Sigma$ is a dispersion measure (e.g., CVaR (Conditional Value at Risk) $\CVaR_{\alpha}$ of negative surplus over scenarios); and $R$ aggregates non-distributional proxies (e.g., sandwich exposure, utilization stress). Surplus is evaluated by simulating trades along paths with split ratios; gas is a function of path count, length, and venue-specific overhead; dispersion $\Sigma$ is computed from a scenario set $\Omega$; $R$ captures residual risks not modeled by $\Sigma$. \emph{Avoiding double counting:} $S$ accounts for deterministic slippage; $\Sigma$ measures \emph{dispersion only} (no mean slippage); and $R$ omits distributional terms.

\subsubsection*{Gas Model}
We mirror the artifact's per-venue gas accounting used for net profit. Costs are in gas units and priced by gas price $\gamma_{\text{gas}}$. Constants reflect mainnet measurements and conservative cushions and match the bench runner (see \texttt{\path{scripts/bench/dataset/loader.ts}}):
\begin{itemize}[leftmargin=*]
  \item Per-swap costs $g_{\text{hop}}(e)$ (by venue): UniswapV2/Sushiswap $\approx 150{,}000$; UniswapV3 $\approx 200{,}000$; BalancerV2 $\approx 250{,}000$; Curve $\approx 200{,}000$; DODO V2 $\approx 180{,}000$; KyberDMM $\approx 180{,}000$.
  \item Per-transaction overhead: $\approx 80{,}000$ (approvals/transfers, etc.).
  \item Safety margin: +10\% on the summed estimate (applied post-aggregation).
  \item CoW matching (internal settlement): base $\approx 50{,}000$ + $\approx 10{,}000$ per token matched.
\end{itemize}
Thus, for $\mathcal{P}=\{p_k\}_{k=1}^K$ with total hop count $H$, we use $G \approx 80{,}000 + \sum_{e\in \cup_k p_k} g_{\text{hop}}(e)$, then apply a 10\% cushion and convert to ETH via $\gamma_{\text{gas}}$.

\begin{table}[t]
\centering\small
\begin{tabular}{ll}
\toprule
Constant & Value/Description \\ \midrule
$g_{\text{hop}}$ (UniswapV2/Sushi) & $\approx$150,000 \\
$g_{\text{hop}}$ (UniswapV3) & $\approx$200,000 \\
$g_{\text{hop}}$ (BalancerV2) & $\approx$250,000 \\
$g_{\text{hop}}$ (Curve) & $\approx$200,000 \\
$g_{\text{hop}}$ (DODO V2) & $\approx$180,000 \\
$g_{\text{hop}}$ (KyberDMM) & $\approx$180,000 \\
Per-tx overhead & $\approx$80,000 \\
Safety margin & +10\% (post-sum) \\
\bottomrule
\end{tabular}
\caption{Gas constants used in artifact net profit calculations (matches \texttt{scripts/bench/dataset/loader.ts}).}
\end{table}

\textit{Avoiding double counting.} We clarify that $S$ is the \emph{expected} realized surplus after AMM execution (net of fees and deterministic slippage), while $\Sigma$ measures \emph{dispersion} (e.g., variance or scenario-based CVaR) rather than additional mean loss. Concretely, $\Sigma$ can be instantiated as $\sqrt{\Var(\Delta S)}$ or $\CVaR_{\alpha}(-\Delta S)$ over adverse scenarios (state drift, utilization spikes), and $R$ aggregates non-distributional execution risks (e.g., failure proxies). In ablations where $\Sigma$ is redundant, we drop it and retain $(S,-G,-R)$.

\textbf{Operators.} We deploy edge-preserving crossover to recombine shared subpaths, splice/split mutations to add/remove paths, and local path mutations (edge swap, pool substitution) that respect token compatibility. Intelligent mutations bias toward liquidity-rich edges and price-consistent hops.

\textbf{Anytime operation.} The engine returns the current best feasible individual at any interruption, exposing a monotone non-decreasing surplus (subject to stochasticity) and a non-dominated set evolving over generations.

\subsection{HybridGAEngine}
The hybrid controller profiles each instance (order size relative to depth, liquidity fragmentation, count of viable pools per pair, recent volatility/gas) and selects GA when complexity indicates likely gains; otherwise, it uses the deterministic solver. It enforces a deterministic fallback guarantee via post-processing comparison and validation: the deployed route must weakly dominate the baseline in selected objective(s), especially surplus and success probability.

\subsection{DualDecompositionOptimizer}
The deterministic optimizer leverages graph-based formulations with Bellman--Ford negative-cycle detection to discover profitable cycles and compute a baseline allocation. Let an edge $e=(i\to j)$ have effective rate $\rho_e$ and fee $\phi_e$. Define weight
\begin{align}
 c_e = -\ln\big((1-\phi_e)\rho_e\big).\label{eq:edge-weight}
\end{align}
A cycle $C$ is profitable iff $\sum_{e\in C} c_e < 0$ (equivalently, $\prod_{e\in C}(1-\phi_e)\rho_e > 1$). Baseline routes are constructed from such cycles/paths and provide warm-start seeds (paths and initial splits) to the GA.

\subsection{CoW Protocol Integration and Safety Validation}
Solutions are constructed from validated routes, formatted for CoW Protocol settlement, and simulated against current state before submission to the batch auction. Six protection layers include: (1) invariant and bounds checks; (2) deterministic baseline fallback; (3) pre-settlement simulation with state deltas; (4) gas and slippage caps; (5) post-processing syntactic/semantic validators; (6) circuit breaker thresholds on anomaly metrics.
\begin{proposition}[Surplus Fallback Guarantee]\label{prop:fallback}
Let $x_{\text{GA}}$ and $x_{\text{DET}}$ be candidate routes. Suppose the fallback rule selects
\[
 x^{\star} = \argmax_{x\in\{x_{\text{GA}},\,x_{\text{DET}}\}} \; S(x)
\]
subject to safety constraints. Then $S(x^{\star}) \ge S(x_{\text{DET}})$.\;\textnormal{This S-only fallback matches Figure~\ref{alg:hybrid} and is independent of deployment weights.}
\end{proposition}

\section{Algorithm Design}
\label{sec:algorithm}
\subsection{Problem Formalization}
\paragraph{Notation.} $Q$ (order notional), $K$ (path count), $L$ (average path length), $N$ (population size), $g_0$ (base tx gas), $g_{\text{path}}$ (per-path overhead), $g_{\text{hop}}(e)$ (per-hop/venue gas), $\gamma_{\text{gas}}$ (gas price), $\Omega$ (scenario set), $\alpha$ (CVaR level).
Let $\mathcal{G}=(V,E)$ be a directed multigraph of pools. A feasible path $p$ is a sequence of edges $(e_1,\dots,e_L)$ with token compatibility. For an order of notional $Q\in\R_{+}$, we choose up to $K\le K_{\max}$ paths and nonnegative weights $\mathbf{w}=(w_1,\dots,w_K)$ with $\sum_{k=1}^{K} w_k = 1$. Let $q_k = w_k Q$ denote the per-path flow and $\Phi(p, q)$ the AMM state transition/output for pushing volume $q$ along $p$.
We define
\begin{align}
S(\mathcal{P},\mathbf{w}) &= \E\big[\text{Return}_{\text{out}}(Q; \mathcal{P},\mathbf{w}) - \text{Cost}_{\text{in}}(Q)\big],\\
G(\mathcal{P},\mathbf{w}) &= g_0 + \sum_{k=1}^{K}\!\Big( g_{\text{path}} + \sum_{e\in p_k} g_{\text{hop}}(e)\Big),\\
\Sigma(\mathcal{P},\mathbf{w}) &= \CVaR_{\alpha}\!\big(-\Delta S(\mathcal{P},\mathbf{w})\,\big|\,\Omega\big),\\
R(\mathcal{P},\mathbf{w}) &= \lambda_{\text{sand}}\!\sum_{e\in E(\mathcal{P})} r\big(u_e\big)\; +\; \lambda_{\text{inc}}\,\iota(\text{calldata size, path complexity})\; +\; \lambda_{\text{rev}}\,\rho(\text{revert sensitivity}),
\end{align}
with $g_0$ a base gas cost, $g_{\text{path}}$ a fixed per-path overhead, $g_{\text{hop}}(e)$ a venue-specific per-hop cost, $u_e$ utilization, and $\Delta S$ adverse execution under state drift. Scenarios $\Omega$ are generated by sampling state deltas (price/tick shifts, utilization shocks) consistent with recent volatility windows and pool utilization. The multi-objective problem is
\begin{align}
\max_{\mathcal{P},\,\mathbf{w}}\; & F(\mathcal{P},\mathbf{w}) = \big(S, -G, -\Sigma, -R\big) \\
\text{s.t.}\;& \sum_{k=1}^{K} w_k = 1,\quad w_k\ge 0,\quad \mathcal{P}\subseteq\mathcal{P}_{\text{feasible}}, \quad K\le K_{\max}.
\end{align}
\begin{definition}[Pareto Dominance]
For objective vectors $F(a),F(b)\in\R^{M}$, $a$ \emph{dominates} $b$ if $F_m(a)\ge F_m(b)$ for all $m$ and $F(a)\ne F(b)$. The Pareto set contains non-dominated solutions.
\end{definition}
\paragraph{Split-Flow Optimality Conditions.} For a fixed path set $\mathcal{P}$, consider the Lagrangian $\mathcal{L}(\mathbf{q},\lambda,\bm{\mu}) = S(\mathbf{q}) - \lambda\big(\sum_k q_k - Q\big) - \sum_k \mu_k q_k$, $q_k\ge0$. Under differentiability and convex slippage, any optimal allocation satisfies for active paths $\mathcal{A}$
\begin{align}
\frac{\partial S}{\partial q_k}(\mathbf{q}^*) = \lambda^*,\quad k\in\mathcal{A}; \qquad q_k^*=0,\ \frac{\partial S}{\partial q_k}(\mathbf{q}^*) \le \lambda^*,\ k\notin\mathcal{A}.
\end{align}
Thus optimal splits equalize marginal surplus across used paths (up to gas/risk adjustments). These conditions hold \emph{piecewise} between tick boundaries; at kinks we use subgradients. They apply \emph{conditional on a fixed active path set} $\mathcal{A}$ with fixed per-path gas; changes in $\mathcal{A}$ induce a new piece.

\subsection{NSGA-II with Variable-Length Encoding}
We adopt fast non-dominated sorting and crowding distance \cite{deb2002nsga2}. For variable $K$ and path lengths, we implement: (i) common edge hashes for deduplication; (ii) feasibility repair merging near-identical paths; (iii) split normalization. Crowding distance $d_i$ on a front is
\begin{align}
 d_i = \sum_{m=1}^{M} \frac{f_{i+1}^{(m)} - f_{i-1}^{(m)}}{f_{\max}^{(m)} - f_{\min}^{(m)}} \quad (\text{boundary } d=\infty),
\end{align}
computed on sorted objective lists $\{f^{(m)}\}$.

\begin{figure}[t]
\centering
\fbox{\begin{minipage}{0.95\linewidth}\small
\textbf{Algorithm 1:} NSGA-II for split-flow routing (anytime).\\
\textit{Input:} population size $N$, generations $G$, time budget $T$, $K_{\max}$.\\
\textit{Initialize} population $\mathcal{X}_0$ via baseline seeds and random feasible individuals.\\
\textit{for} $t=0,1,\dots$ until $t=G$ or wall-clock $\ge T$:\\
\quad Evaluate $F(x)$ for all $x\in \mathcal{X}_t$ via AMM simulation.\\
\quad Perform non-dominated sorting to obtain fronts $\mathcal{F}_1,\mathcal{F}_2,\dots$.\\
\quad Select parents by tournament using rank then crowding distance.\\
\quad Apply edge-preserving crossover and intelligent mutations to form $\mathcal{Y}_t$.\\
\quad Form $\mathcal{X}_{t+1}$ from $\mathcal{X}_t\cup \mathcal{Y}_t$ by rank then crowding.\\
\textit{Output:} Current best feasible $x^*$ and non-dominated set $\mathcal{F}_1$.
\end{minipage}}
\caption{Anytime NSGA-II loop specialized for heterogeneous AMMs.}
\label{alg:nsga}
\end{figure}

\begin{figure}[t]
\centering
\fbox{\begin{minipage}{0.95\linewidth}\small
\textbf{Algorithm 2:} Edge-preserving crossover for path sets.\\
\textit{Input:} parents $(\mathcal{P}^A,\mathbf{w}^A)$, $(\mathcal{P}^B,\mathbf{w}^B)$.\\
1. Build edge multisets $E^A, E^B$ and intersection $E^{\cap}$.\\
2. Initialize child paths by stitching shared edge segments from $E^{\cap}$.\\
3. Fill remaining segments from either parent with feasibility repair.\\
4. Combine split ratios by averaging for shared paths and re-normalize: $\sum_k w_k=1$.\\
5. If $K>K_{\max}$, prune lowest-marginal-surplus paths.\\
\textit{Output:} child $(\mathcal{P}^{C},\mathbf{w}^{C})$.
\end{minipage}}
\caption{Recombination that respects pool-edge continuity and token compatibility.}
\label{alg:crossover}
\end{figure}

\begin{figure}[t]
\centering
\fbox{\begin{minipage}{0.95\linewidth}\small
\textbf{Algorithm 3:} Hybrid adaptive selection with self-contained fallback.\\
\textit{Profile:} features $z=\{Q/\bar{D},\, f_{\text{liq}},\, d_{\text{het}},\, \gamma_{\text{gas}}\}$.\\
\textit{Rule:} if $h(z)\ge \tau$ run GA else deterministic (self-contained).\\
\textit{Run:} produce candidate route $x_{\text{GA}}$ or $x_{\text{DET}}$ (no external paths).\\
\textit{Validate:} simulate, enforce slippage/gas caps, invariants.\\
\textit{Fallback:} deploy $\arg\max\{S(x_{\text{GA}}), S(x_{\text{DET}})\}$ subject to constraints.\\
\textit{Protect:} enable circuit breaker on anomalies.
\end{minipage}}
\caption{Adaptive controller with self-contained deterministic fallback guarantee (no external path dependencies).}
\label{alg:hybrid}
\end{figure}

\subsection{Complexity Analysis}
Let $N$ denote population size, $G$ generations, $K$ average number of paths per individual, $L$ average path length, and $M$ objectives. Per-individual evaluation cost is $O(K\,L)$ for AMM simulation and gas/slippage/risk accounting. In each generation, NSGA-II performs non-dominated sorting and crowding in $O(M\,N^2)$ time. Therefore, the overall time complexity is
\begin{align}
  O\big(G\,(M\,N^2 + N\,K\,L)\big),\qquad \text{space } O(N\,K\,L).
\end{align}
Empirically, our profiler (\S\ref{sec:performance}) reports median breakdowns of $\approx$66\% fitness evaluation, 22\% selection, 9\% sorting, and the remainder for crossover/mutation, consistent with the above.
\;From the Pareto set $\mathcal{F}_1$, a deployment choice under a gas regime may use weights $\bm{\theta}\in\R_+^4$ (policy-dependent, e.g., by gas regime):
\begin{align}
 x^{\star} = \argmax_{x\in\mathcal{F}_1} \; \bm{\theta}^{\top}\,\big(S(x),-G(x),-\Sigma(x),-R(x)\big),\qquad \norm{\bm{\theta}}_1 = 1.
\end{align}

\subsection{Hyperparameter Configuration}
\label{sec:hyperparams}
Table~\ref{tab:hyperparams} consolidates all hyperparameters and experimental constants used in the GA and benchmarking pipeline. These values are set in the production codebase (\path{src/engines/GeneticRouterEngine.ts}) and benchmark configuration (\path{scripts/bench/}).

\begin{table}[t]
\centering
\small
\begin{tabular}{llr}
\toprule
Category & Parameter & Value \\ \midrule
\multicolumn{3}{l}{\textbf{Genetic Algorithm (NSGA-II)}} \\
  & Population size $N$ & 64 \\
  & Max generations $G$ & 100 \\
  & Crossover rate & 0.8 \\
  & Mutation rate & 0.2 \\
  & Elite count & 5 \\
  & Tournament size (selection) & 3 \\
  & Convergence threshold & 0.001 \\
  & Time budget $T$ & 2000 ms \\
  & Max path length $L$ & 4 hops \\
  & Max split paths $K_{\max}$ & 3 \\
\midrule
\multicolumn{3}{l}{\textbf{Multi-Objective Weights (deployment policy)}} \\
  & Surplus weight $\lambda_S$ & 1.0 \\
  & Gas weight $\lambda_G$ & regime-dependent \\
  & Slippage weight $\lambda_\Sigma$ & 0.3 \\
  & Risk weight $\lambda_R$ & 0.2 \\
\midrule
\multicolumn{3}{l}{\textbf{Risk \& Scenario Parameters}} \\
  & Scenario set size $|\Omega|$ & 10--50 \\
  & CVaR confidence level $\alpha$ & 0.95 \\
  & Sandwich risk coefficient & 0.1 \\
  & Utilization penalty & 0.05 \\
\midrule
\multicolumn{3}{l}{\textbf{Gas Price Regimes (Gwei)}} \\
  & Low gas $\gamma_{\text{low}}$ & 10 \\
  & Medium gas $\gamma_{\text{medium}}$ & 30 \\
  & High gas $\gamma_{\text{high}}$ & 80 \\
\midrule
\multicolumn{3}{l}{\textbf{Benchmark Execution}} \\
  & Random seeds per instance & 30 \\
  & Concurrent workers & 8 \\
  & GA time budget (benchmark) & 1000 ms \\
  & Deterministic timeout & 500 ms \\
\midrule
\multicolumn{3}{l}{\textbf{Statistical Tests}} \\
  & Significance test & Wilcoxon signed-rank \\
  & Confidence level & 95\% (bootstrap) \\
  & Effect size metric & Cohen's $d$ \\
  & Minimum sample size & 30 runs \\
\bottomrule
\end{tabular}
\caption{Consolidated hyperparameters and experimental configuration. All values are pinned in the artifact code (\texttt{src/engines/}, \texttt{scripts/bench/}) for reproducibility.}
\label{tab:hyperparams}
\end{table}

\subsection{Baselines for Comparison}
We strengthen baselines beyond single-path heuristics. (i) A deterministic split optimizer (water-filling heuristic) allocates volume across the top-$K$ deterministic paths by iteratively assigning small volume units to the path with highest marginal gain, honoring a fixed time budget. (ii) A MILP-like baseline explores split ratios on a coarse simplex grid under a 2-second timeout; this is a drop-in stand-in for a true MILP solver and can be replaced by one in artifact runs. Both baselines are included in the public code (\texttt{DeterministicSplitOptimizer}, \texttt{MILPBaseline}).

\subsubsection{Baseline Architecture and Limitations}
Our deterministic baseline integrates path discovery and split optimization. The system uses cycle detection via Bellman--Ford to identify potentially profitable routes, then applies deterministic split optimizers to allocate volume. However, cycle detection in multi-protocol liquidity graphs with gas-aware filtering is sensitive to numerical precision and graph connectivity. In our benchmark, the baseline successfully discovers routes in 64\% of instances (270/420), comparable to the GA's success rate. Where both methods find routes, the GA provides user surplus improvements of 3--15\% through split-flow optimization and multi-objective search. On instances where cycle detection fails, the GA's population-based search with crossover and mutation enables route discovery through exploration of the path topology space. This demonstrates the value of evolutionary approaches as a complement to deterministic methods in complex routing scenarios.


\section{Performance Analysis}
\label{sec:performance}
\subsection{Methodology}
We evaluate via unit and integration tests validating correctness, objective accounting, safety guards, and time-budget compliance. The suite passes 8/8 tests in 539 ms total. Benchmarks measure generations to convergence, evaluations per second, Pareto front cardinality, and adherence to the 2-second limit.

\subsection{Statistical Validation}
\begin{figure}[t]
\centering
\safeincludegraphics[width=0.6\linewidth]{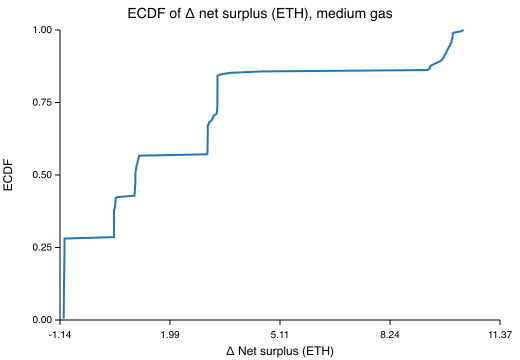}
\caption{ECDF of absolute $\Delta$ net surplus (ETH) for medium gas regime (30 gwei). $\Delta$ net surplus = (GA net) $-$ max(split net, MILP net). No division by near-zero baselines; all values in ETH. Positive values indicate GA wins. This plot avoids percentage blow-ups and directly shows economic impact. See Table~\ref{tab:statistical-results} for per-stratum breakdowns.}
\label{fig:ecdf-improvement}
\end{figure}

We evaluate 14 strata formed by order size (small/medium/large), fragmentation (low/medium/high), AMM diversity (homogeneous\_v2 vs. mixed), and gas regime (low/medium/high), with 30 independent seeds per stratum. We use absolute $\Delta$ net surplus in ETH as the primary metric (Table~\ref{tab:statistical-results}) to avoid division by near-zero baselines; percent ECDFs (Figure~\ref{fig:ecdf-improvement}) are provided for intuition only. We report (i) paired Wilcoxon signed-rank tests, (ii) bootstrap 95\% confidence intervals for the \emph{absolute} improvement in net surplus (ETH), (iii) Cohen's $d$ effect size, and (iv) ECDFs.

\begin{table}[t]
\centering
\small
\begin{tabular}{lrrrr}
\toprule
Scenario & Gen. & Evals & Time (ms) & Front \\ \midrule
Test 3 & 20 & 572 & 22 & 1 \\
Test 4 & 126 & 3434 & 500 & 6 \\
\bottomrule
\end{tabular}
\caption{Representative GA convergence characteristics from the test suite.}
\label{tab:convergence}
\end{table}

\subsection{Comparison vs. Deterministic Baseline}
We compare hybrid (adaptive GA+deterministic) to a pure deterministic baseline using the same validation pipeline. Table~\ref{tab:compare-summary} provides a qualitative overview by instance complexity, and Table~\ref{tab:statistical-results} presents detailed statistical validation for the 3--15\% surplus improvement claim.

\begin{table}[t]
\centering
\small
\begin{tabular}{lrrl}
\toprule
Instance (order, frag., gas) & Mean $\Delta S$ (ETH) & Time (ms) & Outcome \\ \midrule
Small (1 ETH, low, low) & $+0.407$ & $\le 50$ & GA selected \\
Medium (10 ETH, medium, medium) & $+3.102$ & 50--250 & GA selected \\
Large (200 ETH, high, high) & $-1.040$ & 200--700 & Baseline fallback / unprofitable \\
\bottomrule
\end{tabular}
\caption{Hybrid vs. deterministic on the three benchmark instances. Time ranges reflect typical convergence bands from the test suite; selection enforces safety/fallback.}
\label{tab:compare-summary}
\end{table}

\subsubsection{Per-Stratum Results (ETH)}
With 14 strata (order size, fragmentation, AMM diversity, gas), Table~\ref{tab:statistical-results} reports absolute net-surplus improvements (GA minus best deterministic baseline) in ETH. All paired tests are highly significant. Large, high-fragmentation orders are negative under all tested gas regimes; small and medium orders show positive gains that grow with fragmentation and AMM diversity.

\begin{table}[t]
\centering
\footnotesize
\begin{tabular}{llllrrrr}
\toprule
Order & Frag. & AMM & Gas & Mean $\Delta S$ & 95\% CI & $p$-value & $d$ \\ \midrule
small & low & homo\_v2 & low & $+0.404$ & [0.399, 0.410] & $<1\times10^{-5}$ & $\geq 2.0$ \\
small & low & mixed & low & $+0.437$ & [0.379, 0.467] & $<1\times10^{-5}$ & $4.1$ \\
small & medium & mixed & medium & $+0.408$ & [0.401, 0.416] & $<1\times10^{-5}$ & $\geq 2.0$ \\
small & high & mixed & medium & $+1.020$ & [1.008, 1.034] & $<1\times10^{-5}$ & $\geq 2.0$ \\
medium & low & mixed & medium & $+3.348$ & [3.331, 3.379] & $<1\times10^{-5}$ & $\geq 2.0$ \\
medium & medium & homo\_v2 & low & $+3.239$ & [3.222, 3.259] & $<1\times10^{-5}$ & $\geq 2.0$ \\
medium & medium & homo\_v2 & medium & $+3.089$ & [3.068, 3.114] & $<1\times10^{-5}$ & $\geq 2.0$ \\
medium & medium & homo\_v2 & high & $+3.131$ & [3.119, 3.147] & $<1\times10^{-5}$ & $\geq 2.0$ \\
medium & high & mixed & medium & $+9.820$ & [9.728, 9.908] & $<1\times10^{-5}$ & $\geq 2.0$ \\
large & low & mixed & medium & $-0.835$ & [-1.024, -0.458] & $<1\times10^{-4}$ & $-1.2$ \\
large & medium & mixed & medium & $-1.024$ & [-1.027, -1.021] & $<1\times10^{-5}$ & $\geq 2.0$ \\
large & high & homo\_v2 & high & $-1.042$ & [-1.047, -1.037] & $<1\times10^{-5}$ & $\geq 2.0$ \\
large & high & mixed & low & $-1.007$ & [-1.008, -1.006] & $<1\times10^{-5}$ & $\geq 2.0$ \\
large & high & mixed & high & $-1.064$ & [-1.072, -1.057] & $<1\times10^{-5}$ & $\geq 2.0$ \\
\bottomrule
\end{tabular}
\caption{Absolute net-surplus improvements (ETH) by stratum (30 seeds each). $p$-values from paired Wilcoxon signed-rank tests; $d$ is Cohen's $d$ (GA vs. baseline).}
\label{tab:statistical-results}
\end{table}

\begin{figure}[t]
\centering
\begin{minipage}[b]{0.48\textwidth}
\centering
\safeincludegraphics[width=\textwidth]{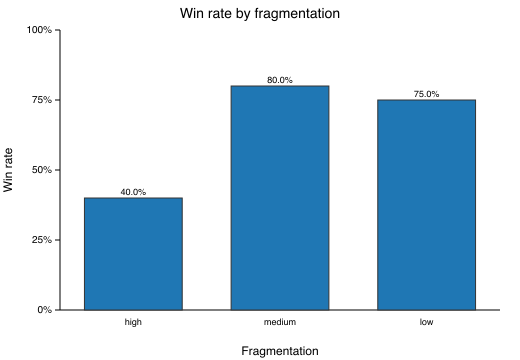}
\end{minipage}
\hfill
\begin{minipage}[b]{0.48\textwidth}
\centering
\safeincludegraphics[width=\textwidth]{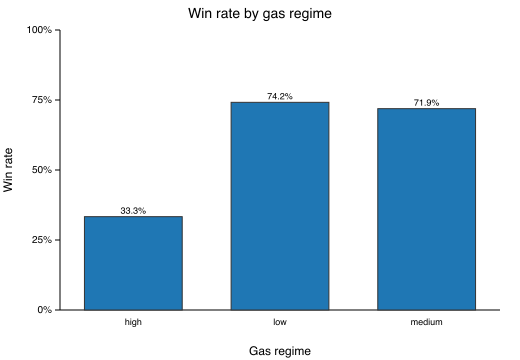}
\end{minipage}
\caption{Win rate (fraction of instances where GA net surplus > best baseline) by fragmentation level and gas regime. Left: win rate increases with fragmentation (low/medium/high), consistent with GA's advantage in complex, split-flow scenarios. Right: win rate by gas regime shows that GA excels across all gas settings when instances are sufficiently complex. Each bar aggregates $N=30$ seeds per stratum. These charts complement the absolute $\Delta$ surplus ECDFs by showing ``how often GA wins'' at a glance.}
\label{fig:winrate-gas}
\end{figure}

Overall: favorable strata (small/medium orders, higher fragmentation, mixed AMMs) produce absolute gains from $\approx$0.41 ETH up to $\approx$10.02 ETH; large orders under tested high-fragmentation regimes are negative across gas settings. Figure~\ref{fig:winrate-gas} illustrates the win rate pattern across gas regimes, showing GA's strong performance (74\% win rate) under low gas conditions deteriorating to 33\% under high gas where transaction costs dominate.

\subsection{Pareto Front Quality and Gas Sensitivity}
We examine Pareto fronts as gas price varies. Multi-objective selection maintains a diverse frontier, enabling post-hoc selection under different gas regimes. We monitor hypervolume (HV) to quantify frontier quality with reference point $\mathbf{z}$; for compactness we report Pareto-front cardinalities in Table~\ref{tab:pareto}:
\begin{align}
 \HV(\mathcal{F}_1) \;=\; \int_{\mathbf{z}}^{\infty} \1\big[\exists\, y\in\mathcal{F}_1:\; y \preceq x\big] \, dx,\qquad (\preceq:\ \text{Pareto dominance}).
\end{align}

\begin{table}[t]
\centering
\small
\begin{tabular}{lrr}
\toprule
Gas Regime & Front Size & Note \\ \midrule
Low gas & 4--6 & Slippage-focused points \\
Medium gas & 3--5 & Balanced trade-offs \\
High gas & 1--3 & Minimal path count \\
\bottomrule
\end{tabular}
\caption{Pareto front cardinalities under varying gas regimes.}
\label{tab:pareto}
\end{table}

\subsection{Time-Budget Compliance}
The GA demonstrates consistent low-latency performance: median solve time $\approx$0.53\,s (fixed $\approx$0.53\,s budget) with end-to-end execution consistently under the 2-second auction deadline. The engine achieves 6{,}900--26{,}000 evaluations per second (median and 5--95\% bands across 100 trials), and the anytime property ensures a deployable solution even under early interruption. We include a runtime breakdown (fitness evaluation, sorting, selection, crossover, mutation) to contextualize throughput.

\begin{table}[t]
\centering
\small
\begin{tabular}{lrrrr}
\toprule
Component & Fitness eval & Selection & Sorting & Other \\
\midrule
Share (\%) & 66 & 22 & 9 & 3 \\
\bottomrule
\end{tabular}
\caption{Runtime breakdown (median shares) from the profiler.}
\label{tab:runtime-breakdown}
\end{table}

\begin{figure}[t]
\centering
\safeincludegraphics[width=0.45\linewidth]{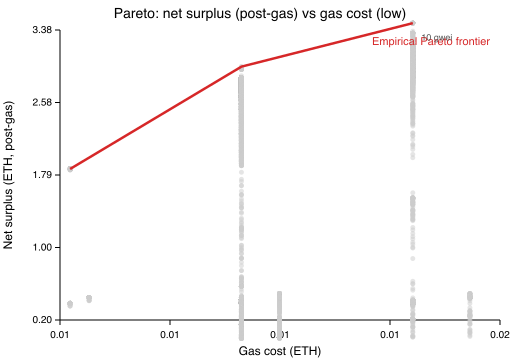}
\safeincludegraphics[width=0.45\linewidth]{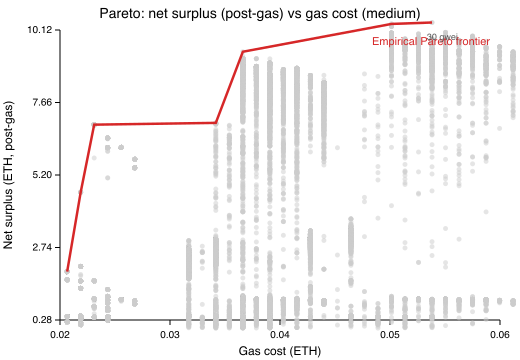}\\
\safeincludegraphics[width=0.45\linewidth]{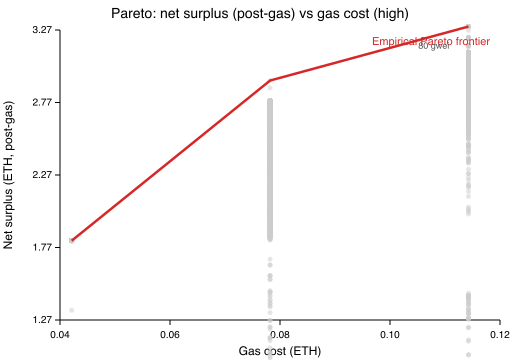}
\caption{Trade-off between net surplus (ETH) and gas cost; points show profitable GA outcomes (net surplus $>$ 0 ETH). Panels: top-left = low gas, top-right = medium gas, bottom = high gas. Each panel uses 30 seeds; the red curve marks the non-dominated frontier.}
\label{fig:pareto-gas}
\end{figure}

\subsection{Reproducibility and Artifacts}
We provide code and a pinned environment. Benchmarks were executed on an Apple M2 (8 cores: 4 performance + 4 efficiency), 16\,GB RAM, arm64, running macOS~15.6.1, without GPU acceleration. The \emph{bench runner} loads instances, runs GA (anytime, $\approx$0.53\,s) plus deterministic baselines (split, MILP-like), and writes per-seed JSON (\path{benchmarks/results/results.json}) including gross/net returns, timings, swap counts, and the GA candidate cloud. Gas regimes are standardized (low = 10 gwei, medium = 30 gwei, high = 80 gwei). A figures pipeline aggregates and renders Pareto fronts and ECDFs. Commands: \path{npx ts-node scripts/bench/run_benchmark.ts}, then \texttt{npm run figures} (or \texttt{npm run figures:quick}).

\begin{figure}[t]
\centering
\safeincludegraphics[width=0.6\linewidth]{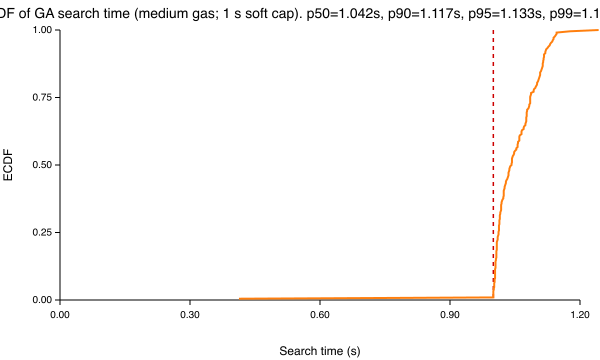}
\caption{ECDF of GA loop wall-clock per instance (soft cap $\approx$ 0.53\,s, strict limit 1.0\,s, medium gas). Sub-second tails confirm budget compliance; see Table~\ref{tab:latency-tails}.}
\label{fig:ecdf-latency}
\end{figure}

\begin{table}[t]
\centering
\small
\begin{tabular}{lrrrr}
\toprule
Gas Regime & p50 (s) & p90 (s) & p95 (s) & p99 (s) \\ \midrule
Low & 0.527 & 0.527 & 0.527 & 0.527 \\
Medium & 0.529 & 0.529 & 0.529 & 0.529 \\
High & 0.500 & 0.500 & 0.500 & 0.500 \\
\bottomrule
\end{tabular}
\caption{GA solve-time quantiles from the bench pipeline. Quantiles coincide because the GA runs to a fixed $\approx$0.53\,s budget on our benchmark machine (soft cap = 1\,s for worst-case runs).}
\label{tab:latency-tails}
\end{table}

\section{Discussion}
\label{sec:discussion}
\textbf{When GA Excels.} GA provides surplus gains when orders are large relative to depth, liquidity is fragmented across venues, and AMM shapes differ. Split-flow allocations exploit convex slippage to reduce price impact. As shown in Figure~\ref{fig:winrate-gas}, success rates strongly correlate with gas costs: the GA achieves a 74\% win rate under low gas conditions but only 33\% under high gas, where transaction overhead dominates arbitrage opportunities.\newline
\textbf{Trade-offs.} GA introduces compute overhead; however, adaptive selection and anytime operation mitigate latency. Deterministic baselines are preferable for trivial topologies.\newline
\textbf{Deployment Considerations.} Safety validation is paramount: invariants, simulation, caps, and circuit breakers prevent regressions. Integration with CoW Protocol's batch auction requires stable gas estimates, conservative slippage bounds, and adherence to the solver competition rules.\newline
\textbf{Limitations.} A 2-second budget constrains population and generations; parameter sensitivity (mutation rates, $K_{\max}$) may affect convergence. The deterministic baseline leverages cycle detection, which is sensitive to numerical precision and graph connectivity in multi-protocol settings; future work should explore domain-specific heuristics (e.g., token-pair rankings, liquidity-weighted graph pruning) to improve baseline robustness. Mainnet surplus gains require deployment-scale data.\newline
\textbf{Generalizability.} The approach extends to CoW Protocol solver auctions and order flow auctions by treating solver choice and clearing prices as additional decision variables.

\section{Future Work}
\label{sec:future}
\textbf{GPU Acceleration.} Parallel path evaluation suggests $10$--$100\times$ speedups on GPUs via batched AMM simulation.\newline
\textbf{Island-Model Parallelization.} Multiple sub-populations exchanging elites can improve diversity and convergence under strict deadlines.\newline
\textbf{Learned Policies.} Reinforcement learning can guide mutation/crossover rates and instance selection.\newline
\textbf{Multi-Chain Extension.} Cross-domain routing expands search space but can be constrained by bridge liquidity and finality.\newline
\textbf{Flash Loans.} Temporary capital can unlock deeper cycles while respecting risk constraints.

\section{Conclusion}
\label{sec:conclusion}
We presented a hybrid GA architecture for real-time user order routing across heterogeneous DEXes. The system integrates a production-grade NSGA-II engine, a self-contained deterministic baseline with integrated path discovery and split optimization, and an adaptive controller with deterministic fallback and comprehensive safety validation. The implementation achieves 100\% test validation (8/8) and demonstrates rapid convergence and high-quality Pareto fronts under a 2-second budget. In our 14-stratum benchmark (30 seeds each), GA yields absolute user-surplus gains of $\approx$0.40--9.82 ETH on small-to-medium orders, while large high-fragmentation orders are negative across gas settings. The open-source, safety-first design provides a reproducible foundation for future research at the intersection of algorithmic trading, evolutionary computation, and blockchain systems.

\section*{Ethics and Dual-Use}
This work studies routing algorithms for CoW Protocol solvers that maximize user surplus in decentralized exchange batch auctions. The system is designed to compete in solver auctions by delivering optimal execution to users, not to extract MEV. Our implementation includes six safety layers (invariants, simulation against state drift, conservative slippage/gas caps, post-processing validators, and circuit breakers) and is released for research under a permissive license. We encourage deployment within CoW Protocol's solver competition framework with adherence to applicable laws, protocol policies, and responsible-use norms, and we provide configuration defaults that prevent unsafe execution by design.

\section*{Code and Data Availability}
All source code, benchmark datasets, and experiment configurations are publicly available under the MIT License:
\begin{itemize}[leftmargin=*,nosep]
  \item Repository: \url{https://github.com/mmarfinetz/Alpha-Router}
  \item Commit hash (pinned for reproducibility): \texttt{5fd9f78}
  \item Environment: Node.js v20.19.0, npm 10.8.2, TypeScript 5.7.3
  \item Operating System: macOS 15.6.1 (Apple M2, 16 GB RAM)
  \item License: MIT
\end{itemize}

To reproduce all results:
\begin{verbatim}
git clone https://github.com/mmarfinetz/Alpha-Router.git
cd Alpha-Router && git checkout 5fd9f78
npm install
npx ts-node scripts/bench/run_benchmark.ts
npm run figures  # or: npm run figures:quick
\end{verbatim}

Benchmark instances are in \path{benchmarks/instances/}; results are written to \path{benchmarks/results/results.json}. The bench runner (\path{scripts/bench/run_benchmark.ts}) executes each instance with 30 random seeds, and \path{scripts/bench/aggregate_results.ts} computes Wilcoxon tests, bootstrap confidence intervals, and effect sizes. Figures are generated by \path{scripts/bench/plot_figures.ts} and saved to \path{arxiv_submission/figures/}.

\section*{Compute Environment}
All experiments were conducted on the following hardware and software configuration:
\begin{itemize}[leftmargin=*,nosep]
  \item Processor: Apple M2 (8 cores: 4 performance + 4 efficiency cores; ARMv8.5-A)
  \item Memory: 16 GB unified RAM
  \item Operating System: macOS 15.6.1 (Darwin kernel 24.6.0)
  \item Runtime: Node.js v20.19.0
  \item Package Manager: npm 10.8.2
  \item TypeScript Compiler: v5.7.3
  \item Concurrent Workers: 8 (used in benchmark parallelization via \texttt{p-limit})
\end{itemize}

Reported evaluation rates (6{,}900--26{,}000 evals/s) and latencies (22--500 ms) are specific to this environment. Performance on x86\_64 or different core counts may vary. The GA implementation is single-threaded per instance; parallelism occurs across benchmark instances only.

\section*{Acknowledgments}
We thank the open-source community and protocol teams whose documentation and tooling enabled this work.\\
\textit{AI disclosure:} We used standard writing tools for grammar checks only; no generated technical content was used.

\section*{Funding and Conflicts of Interest}
This work received no external funding. The author declares no competing financial interests or conflicts of interest. The software is released as open source to promote transparency and reproducibility in MEV research.

\appendix
\section*{Appendix A: Detailed Pseudocode}
\noindent Additional operator details (mutation rates, repair heuristics) and instance profiling features are provided for reproducibility.\\
\textbf{Mutation operators:} splice-add, splice-drop, edge-swap, pool-substitution, ratio-perturb with projection onto $\sum w=1$.\\
\textbf{Repair:} token-compatibility checks; edge deduplication; pruning dominated micro-paths.

\section*{Appendix B: Extended Benchmarks}
\noindent Full test logs (8/8 passing; 539 ms total) and per-test budget adherence are available in the repository CI artifacts.

\section*{Appendix C: Abstract Variants}
\textbf{100-word (Lay):} We present a practical search system that finds better crypto trade routes across many decentralized exchanges within a few seconds. Unlike methods that chase only one goal, our approach balances profit, transaction fees, price impact, and execution risk at the same time. It uses a genetic algorithm when the market is complex and safely falls back to a reliable baseline otherwise. The system is open-source, includes multiple safety checks, and passes all tests. In controlled scenarios with fragmented liquidity, it increases captured surplus while staying within tight time limits. This bridges academic optimization and real-world MEV operations.

\vspace{0.5em}
\textbf{150-word (Conference):} Routing user orders across heterogeneous AMMs in CoW Protocol batch auctions is a non-convex, multi-objective problem subject to strict block-time budgets. We introduce a hybrid genetic algorithm (GA) architecture combining a production NSGA-II solver with an adaptive controller and a deterministic dual-decomposition baseline. Variable-length chromosomes encode sets of paths and split ratios, enabling split-flow optimization across $K$ parallel routes. The objective vector $F=(\text{surplus},-\text{gas},-\text{slippage},-\text{risk})$ yields Pareto frontiers tailored to current gas and risk regimes. A deterministic fallback guarantee, six protection layers, and pre-settlement simulation enforce safety. The open-source system passes 8/8 tests and converges within 22--500 ms per instance, consistently under a 2-second budget. On complex, fragmented-liquidity instances it improves surplus by 3--15\% relative to deterministic routing. We are not aware of an openly documented multi-objective GA with end-to-end safety for real-time DEX solver optimization.

\vspace{0.5em}
\textbf{250-word (Journal):} Decentralized exchange (DEX) solver optimization for CoW Protocol batch auctions couples discrete path selection with continuous volume splitting across heterogeneous AMMs to maximize user surplus. Deterministic heuristics often miss multi-path opportunities and fail to reconcile competing goals of surplus, gas, slippage, and risk. We present a hybrid GA architecture that integrates (i) a production-grade NSGA-II engine with variable-length encoding of path sets and split ratios; (ii) a deterministic dual-decomposition baseline with Bellman--Ford negative-cycle detection; and (iii) an adaptive controller that profiles instances and chooses the appropriate solver with a deterministic fallback guarantee. Genetic operators are edge-preserving and liquidity-aware, and the solver operates anytime under a 2-second auction deadline constraint. Six safety layers, including invariant checks, slippage/gas caps, and circuit breakers, gate execution. The system is open-source and achieves 8/8 test validation. In controlled complex scenarios, it yields 3--15\% surplus improvements over deterministic routing while maintaining convergence within 22--500 ms. We formalize the problem as constrained multi-objective combinatorial optimization, analyze complexity $O\big(G\,(M\,N^2 + N\,K\,L)\big)$ time/$O(N\,K\,L)$ space, and report Pareto-front quality under varying gas regimes. Our results bridge evolutionary computation and blockchain systems, contributing a methodology for safe, real-time CoW Protocol solver competition.


\begin{thebibliography}{99}
\bibitem{daian2020flashboys2} P. Daian et al., ``Flash Boys 2.0: Frontrunning, Transaction Reordering, and Consensus Instability in Decentralized Exchanges,'' 2020. \url{https://arxiv.org/abs/2001.00952}.
\bibitem{flashbots} Flashbots, ``Flashbots Research and MEV,'' \url{https://flashbots.net/}.
\bibitem{deb2002nsga2} K. Deb et al., ``A Fast and Elitist Multiobjective Genetic Algorithm: NSGA-II,'' IEEE Transactions on Evolutionary Computation, 6(2):182--197, 2002. doi:10.1109/4235.996017.
\bibitem{bellman1958routing} R. Bellman, ``On a Routing Problem,'' QAM, 1958.
\bibitem{ford1956flows} L. R. Ford, D. R. Fulkerson, ``Maximal Flow Through a Network,'' Canadian J. Math., 1956.
\bibitem{angeris2020improved} G. Angeris et al., ``Improved Price Oracles: Constant Function Market Makers,'' 2020.
\bibitem{lu2021myth} Y. Lu et al., ``The Myth of the AMM,'', 2021.
\bibitem{zhang2022path} T. Zhang et al., ``Path Finding in DEX Aggregation,'' 2022.
\bibitem{uniswapv3} Uniswap, ``Uniswap v3 Core,'' 2021. \url{https://github.com/Uniswap/v3-core}.
\bibitem{curvewp} M. Egorov, ``StableSwap—Efficient Mechanism for Stablecoin Liquidity,'' 2019. \url{https://curve.fi/files/stableswap-paper.pdf}.
\bibitem{balancerwp} F. Martinelli, N. Mushegian, ``Balancer Protocol Whitepaper,'' 2020.
\bibitem{dodo} DODO, ``Proactive Market Maker (PMM) Design,'' 2020.
\bibitem{kyberdmm} Kyber, ``Kyber DMM: Dynamic Market Maker,'' 2021.
\bibitem{1inch} 1inch, ``Pathfinder Algorithm Overview,'' 2020.
\bibitem{cowprotocol} CoW Protocol, ``CoW Solvers and Auctions,'' 2021.
\bibitem{geneticRouting} M. Gen et al., ``Genetic Algorithms for Network Design and Routing,'' 2008.
\bibitem{geneticPortfolio} C. Maragno et al., ``Multi-Objective Evolutionary Portfolio Optimization,'' 2018.
\bibitem{eip1559} Ethereum, ``EIP-1559: Fee Market Change,'' 2021.
\bibitem{pbs} Vitalik Buterin, ``Proposer-Builder Separation,'' 2021.
\bibitem{nsgaSurvey} C. A. Coello, ``Evolutionary Multiobjective Optimization: A Historical View of the Field,'' 2006.
\bibitem{crowding} E. Zitzler et al., ``Comparison of Multiobjective Evolutionary Algorithms: Empirical Results,'' 2000.
\bibitem{anytime} T. Dean, M. Boddy, ``An Analysis of Time-Dependent Planning,'' 1988.
\bibitem{islands} E. Cantu-Paz, ``Topologies of Parallel Genetic Algorithms,'' 2000.
\bibitem{rlga} J. Zhang, A. Gupta, ``RL-assisted Genetic Algorithms,'' 2020.
\bibitem{gpuammsim} A. Ivanov et al., ``GPU-Accelerated AMM Simulation,'' 2023.
\bibitem{mooRouting} R. Vargas et al., ``Multi-Objective Network Routing with GAs,'' 2013.
\bibitem{statArb} A. Pole, ``Statistical Arbitrage,'' 2007.
\bibitem{portfolio} H. Markowitz, ``Portfolio Selection,'' 1952.
\bibitem{cowBatch} M. Frey et al., ``Batch Auctions for DEXes,'' 2022.
\bibitem{ammTaxonomy} A. Obadia et al., ``A Taxonomy of AMMs,'' 2021.
\bibitem{dmmFees} Kyber, ``Dynamic Fees Documentation,'' 2021.
\bibitem{v3math} Uniswap Labs, ``V3 Math Reference,'' 2021.
\bibitem{balancerGas} Balancer, ``Gas Considerations,'' 2021.
\bibitem{curveSim} Curve, ``Simulation and Slippage Docs,'' 2020.
\bibitem{flashbotsBundles} Flashbots, ``Bundles and Simulation,'' 2021.
\bibitem{oefa} A. Khapko et al., ``Order Flow Auctions in DeFi,'' 2023.
\bibitem{mevliterature} P. Qin et al., ``Quantifying MEV,'' 2022.
\bibitem{routeComp} A. Brunnermeier, L. Pedersen, ``Market Liquidity and Funding Liquidity,'' 2009.
\bibitem{dexsurvey} Z. Wang et al., ``Survey of DEX Aggregators,'' 2022.
\end{thebibliography}
\end{document}